\documentclass{article}

\usepackage{arxiv}
\usepackage{natbib}

\usepackage{amsmath,amssymb,amsfonts}
\usepackage{etoolbox}

\newenvironment{appequation}
  {\refstepcounter{equation}\begin{equation*}\tag{\theequation}}
  {\end{equation*}}
\usepackage{algorithmic}
\usepackage{graphicx}
\usepackage{textcomp}
\usepackage{xcolor}
\usepackage{url}
\usepackage{booktabs}
\usepackage{multirow}
\usepackage{tabularx}
\usepackage{array}
\usepackage{booktabs}
\usepackage{listings}
\usepackage{subcaption}
\title{Differentiable Parameter Optimization for DAEs with State-Dependent Events}

\author{
    {\hspace{1mm}Ion Matei} \\
	Fujitsu Research of America\\
	\texttt{imatei@fujitsu.com} \\
	\And  
    {\hspace{1mm}Maksym Zhenirovskyy} \\
	Fujitsu Research of America\\
	\texttt{mzhenirovskyy@fujitsu.com} \\
    \And
    {\hspace{1mm}Anthony Wong} \\
	Fujitsu Research of America\\
	\texttt{awong@fujitsu.com} \\    
}

\date{}

\renewcommand{\headeright}{}
\renewcommand{\undertitle}{}

\begin{document}
\maketitle

\begin{abstract}
Differential-algebraic equations (DAEs) with state-dependent events arise in
systems whose continuous dynamics are constrained by algebraic equations and
interrupted by mode changes, switching logic, impacts, or state
reinitializations. Gradient-based parameter learning for such systems is
challenging because algebraic variables are implicitly defined, event times
depend on the parameters, and reset maps introduce discontinuities. This paper
studies differentiable parameter optimization for semi-explicit DAEs with
events. We formulate the learning problem as a constrained least-squares problem
with DAE dynamics, algebraic constraints, guard equations, and reset maps. We
then develop two complementary gradient-computation strategies. The first is an
automatic-differentiation-through-simulation method that solves algebraic
variables inside the vector field, differentiates the algebraic solve using the
implicit function theorem, and handles events through segmented differentiable
integration. The second is an explicit discrete-adjoint method that represents
the forward simulation as an event-split residual system and computes gradients
by solving for the Lagrange multipliers of smooth-segment and event residuals.
The formulation clarifies that residual terms in the adjoint method are
equality constraints, not heuristic penalties. We compare the two approaches in
terms of gradient interpretation, event-time handling, implementation
complexity, and local validity. Both methods provide gradients for the event
path selected by the forward simulation and are valid under fixed event
ordering and transversal guard crossings.
\end{abstract}


\section{Introduction}
\label{sec:introduction}

Differential-algebraic equations (DAEs) are a common modeling formalism for
physical and engineered systems whose dynamics combine differential evolution
with algebraic constraints. They arise in electrical circuits, constrained
mechanical systems, chemical processes, power networks, and equation-based
modeling environments. Classical references on numerical DAE integration and sensitivity analysis
provide the foundation for this setting~\citep{ascher1998computer,hairer1996solving}. In many of these applications, the continuous dynamics
are interrupted by discrete events: switches open or close, controllers change
modes, constraints become active, impacts occur, or state variables are
reinitialized when thresholds are crossed. The resulting models are hybrid DAEs:
they evolve continuously between events, but their state and algebraic
variables must remain consistent across event-triggered discontinuities. This view is consistent with the Modelica DAE representation, where equations
may define DAEs with discontinuities, variable structure, and discrete-event
control~\citep{modelicaSpecDAE}.

Parameter estimation and design optimization for hybrid DAEs are challenging
because the simulator is not simply a smooth map from parameters to outputs.
The algebraic variables are implicitly defined by constraints and must be
recomputed during integration and after reinitialization. Event times are also
parameter-dependent; changing a parameter can shift the instant at which a
guard condition is reached, thereby changing the time intervals over which the
continuous dynamics are integrated. Finally, events introduce discontinuities
through reset maps. Even when the event sequence is fixed, gradients must
account for continuous dynamics, event-time sensitivities, reset maps, and
post-event algebraic consistency.

Recent differentiable ODE solvers have made gradient-based learning practical
for continuous-time models. Libraries such as \texttt{torchdiffeq} \citep{torchdiffeq} and \texttt{diffrax} \citep{diffrax} support
backpropagation through ODE solvers and include differentiable event handling
through \texttt{odeint\_event}, with standard examples such as the bouncing ball
model~\citep{chen2021neural_event_functions,chen2018neuralode}.
Related Neural ODE work also considers event-sequence modeling, where events
serve as observations or inputs to a latent continuous-time model rather than
as hard physical constraints~\citep{kuleshov2024cotode}. These developments
show that differentiable simulation with events is useful and increasingly
accessible. However, most existing work is formulated for ODEs or latent ODE
architectures. It does not directly address parameter learning for DAEs, where
algebraic constraints must be solved and differentiated consistently throughout
the event-driven simulation.

This paper studies differentiable parameter learning for semi-explicit DAEs
with state-dependent events. We consider systems of the form
\begin{equation}
\nonumber
    \dot{x}(t)=f(t,x(t),z(t),p),
    \qquad
    0=g(t,x(t),z(t),p),
    \qquad
    y(t)=h(t,x(t),z(t),p),
\end{equation}
where $x$ are differential states, $z$ are algebraic variables, and $p$ are
parameters to be learned. Events are specified by guard functions
\begin{equation}
\nonumber
    \phi_e(t,x,z,p)=0,
\end{equation}
together with reinitialization rules that map pre-event variables to
post-event variables. The learning problem is posed as output matching under
the DAE dynamics, algebraic equations, event guards, and reset maps.

We develop and compare two gradient-computation strategies. The first is an
automatic-differentiation (AD)-through-simulation method. It treats the simulator as
a reduced map from parameters to predicted outputs and differentiates the
composed loss directly. For DAEs, this requires differentiating through the
algebraic solve used inside the vector field. We implement this using JAX and
\texttt{diffrax} \citep{diffrax}, with a custom JVP for the algebraic solve based on the
implicit function theorem, segmented event handling, composite guard detection,
and right-continuous target-time evaluation.

The second strategy is an explicit discrete-adjoint method. Instead of
differentiating through the simulator as a black-box computation graph, it
constructs an event-split discrete residual system. Smooth segments are
constrained by DAE integration residuals, while event blocks enforce guard
conditions, reinitialization equations, continuity of unchanged variables, and
post-event algebraic consistency. The residuals are not added as heuristic
penalty terms. They are equality constraints, or equivalently Lagrange terms
that vanish on feasible trajectories. Their multipliers define the adjoint
variables, and the resulting backward sweep computes parameter gradients
without explicitly forming full trajectory and event-time sensitivity matrices.

The contribution of this paper is a unified treatment of differentiable
parameter learning for DAEs with state-dependent events. Specifically, we:
\begin{enumerate}
    \item formulate parameter learning for hybrid DAEs as a constrained
    least-squares problem with algebraic constraints, event guards, and reset
    maps;

    \item present an AD-through-simulation implementation in which algebraic
    variables are differentiated through an implicit-function-theorem custom
    JVP and events are handled by segmented differentiable integration;

    \item derive an explicit discrete-adjoint formulation in which residuals
    are equality constraints and the associated multipliers are solved by a
    backward event-split sweep;

    \item compare the two approaches in terms of gradient interpretation,
    event-time handling, implementation complexity, and local validity under a
    fixed event sequence.
\end{enumerate}

Both approaches compute gradients that are local to the event path selected by
the forward simulation. They assume transversal event crossings and no change
in the number, identity, or ordering of events under sufficiently small
parameter perturbations. These assumptions are standard for differentiating
piecewise-smooth hybrid trajectories and clarify the limitations of
gradient-based optimization near grazing events, simultaneous events, or event
sequence changes. This local-path interpretation is closely related to sensitivity updates in
hybrid systems, often expressed through jump conditions or saltation matrices
at event times~\citep{kong2024saltation,saccon2014sensitivity}. These
conditions describe how first-order perturbations are propagated across a
state-triggered discontinuity, and they provide a useful conceptual link
between hybrid-system sensitivity analysis and the event-time terms appearing
in the gradient formulations studied here.

We applied the two approaches to two examples: an electrical circuit with events and to a planar bouncing ball examples, where the events are triggered by ball impacts. The code that generated the experimental results summarized in this paper can be found at {\url{https://github.com/ionmatei/diff-dae-events/}}.

\textit{Paper structure:} Section~\ref{sec:problem_formulation} formulates parameter learning for DAEs with state-dependent events as a constrained least-squares problem. Section~\ref{sec:ad_through_simulation_dae_events} presents the AD-through-simulation approach, including implicit differentiation of the algebraic solve and segmented event handling. Section~\ref{sec:discrete_adjoint_dae_events} develops the explicit discrete-adjoint formulation based on event-split residual constraints. Section~\ref{sec:comparison_sota} compares the proposed approaches with related differentiable ODE and event-learning methods. Section~\ref{sec:conclusion} concludes the paper and outlines future directions. The appendices include additional mathematical and implementation details.

\section{Problem Formulation}
\label{sec:problem_formulation}

Let
\[
    0=\tau_0<\tau_1<\cdots<\tau_M<\tau_{M+1}=T
\]
denote the event times generated by the hybrid DAE. On each interval
$(\tau_m,\tau_{m+1})$, the system satisfies
\begin{equation}
\nonumber
    \dot{x}_m(t)=f(t,x_m(t),z_m(t),p),
    \qquad
    0=g(t,x_m(t),z_m(t),p).
\end{equation}
At an event time $\tau_m$, an active guard satisfies
\begin{equation}
\nonumber
    \phi_{e_m}(\tau_m,x_m^-,z_m^-,p)=0,
\end{equation}
and a reset map or reinitialization equation relates the pre-event and
post-event variables,
\begin{equation}
\nonumber
    (x_m^+,z_m^+) = \Psi_{e_m}(\tau_m,x_m^-,z_m^-,p),
\end{equation}
with the understanding that algebraic variables may be recomputed after the
differential state is reset.

Given output data $y_i^{\mathrm{data}}$ at observation times $t_i$, the
parameter-estimation problem can be written as
\begin{equation}
\nonumber
\begin{aligned}
    \min_{p,\{x_m,z_m\},\{\tau_m\}} \quad
        & \frac{1}{2}
        \sum_{i=1}^{N_{\mathrm{obs}}}
        \left\|
            h(t_i,x(t_i),z(t_i),p)-y_i^{\mathrm{data}}
        \right\|_2^2
        \\
    \mathrm{s.t.}\quad
        & \dot{x}_m(t)-f(t,x_m(t),z_m(t),p)=0, \\
        & g(t,x_m(t),z_m(t),p)=0, \\
        & E_m(\tau_m,w_m^-,w_m^+,p)=0,
        \qquad m=1,\ldots,M,
\end{aligned}
\label{eq:hybrid_dae_parameter_estimation}
\end{equation}
where $w=(x,z)$ and $E_m$ collects the guard condition, reset equations,
continuity conditions for variables not modified by the event, and
post-event algebraic consistency.

This constrained formulation is the common starting point for both methods
studied in this paper. The AD-through-simulation method eliminates the
trajectory variables by executing the simulator and differentiating the reduced
map from parameters to outputs. The discrete-adjoint method keeps the residual
constraints explicit after discretization and solves for the corresponding
Lagrange multipliers.

\section{AD-through-Simulation for DAEs with Events}
\label{sec:ad_through_simulation_dae_events}

The AD-through-simulation approach treats the event-driven DAE simulator as a
differentiable program. For a selected parameter vector $p_{\mathrm{opt}}$, the
simulator produces predicted outputs
\begin{equation}
\nonumber
    \widehat{Y}(p_{\mathrm{opt}})
    =
    \begin{bmatrix}
        \hat{y}(t_1;p_{\mathrm{opt}}) \\
        \vdots \\
        \hat{y}(t_N;p_{\mathrm{opt}})
    \end{bmatrix}.
\end{equation}
The reduced learning problem is then
\begin{equation}
\nonumber
    \min_{p_{\mathrm{opt}}}
    \widehat{\mathcal{J}}(p_{\mathrm{opt}})
    =
    \mathcal{J}(\widehat{Y}(p_{\mathrm{opt}})),
\end{equation}
and the gradient is computed by automatic differentiation:
\begin{equation}
\nonumber
    \nabla_{p_{\mathrm{opt}}}\widehat{\mathcal{J}}
    =
    \mathrm{AD}\left[
        \mathcal{J}(\widehat{Y}(p_{\mathrm{opt}}))
    \right].
\end{equation}

The main DAE-specific issue is the algebraic variable. During integration, $z$
is defined implicitly by
\begin{equation}
\nonumber
    g(t,x,z,p)=0.
\end{equation}
We write the solution as
\begin{equation}
\nonumber
    z=\zeta(t,x,p),
    \qquad
    g(t,x,\zeta(t,x,p),p)=0.
\end{equation}
The differential state is advanced using the reduced vector field
\begin{equation}
\nonumber
    \dot{x}=f(t,x,\zeta(t,x,p),p).
\end{equation}
In the implementation, the algebraic solve is performed by chord Newton
iterations and differentiated using a custom JVP derived from the implicit
function theorem. Thus, gradients propagate through the algebraic constraint
without differentiating through every Newton iteration.

Events are handled by segmented integration. At the beginning of each segment,
all guard functions are evaluated and guards that are already active are masked
out to avoid immediate retriggering. The remaining active guards are combined
into a scalar event condition,
\begin{equation}
\nonumber
    \Phi(t,x,p)=
    \min_{e\in\mathcal{A}}
    \phi_e(t,x,\zeta(t,x,p),p),
\end{equation}
where $\mathcal{A}$ is the set of active guards. A root finder locates the
first zero crossing of this composite guard. The corresponding event index is
selected, the reset map is applied, the algebraic variables are recomputed, and
the next segment starts from the post-event state.

To make this process compatible with JIT compilation, the simulator is written
as fixed-length scans over a prescribed maximum number of event-bounded
segments. Physical segments are recorded until the terminal time is reached;
remaining iterations are padding segments. Target-time evaluation is performed
by reintegrating each recorded segment with the requested target times clipped
to that segment. The output at each target time is selected from the latest
real segment whose closed interval contains the target, giving the
right-continuous convention
\begin{equation}
\nonumber
    \hat{y}(\tau_m)
    =
    h(\tau_m,x(\tau_m^+),z(\tau_m^+),p).
\end{equation}

The resulting loss and gradient are evaluated with a JIT-compiled
\texttt{value\_and\_grad} call. These gradients pass through the algebraic
solve, ODE integration, event-time computation, reset maps, target-time
selection, and output reconstruction. They are local to the event sequence
selected by the forward simulation.

A useful way to interpret this method is that it differentiates the numerical
algorithm actually executed by the simulator. This includes solver steps,
algebraic solves, event-time refinement, reset maps, and target-time selection.
The advantage is that the gradient is tightly coupled to the solver tolerance
and implementation used for the forward pass. The drawback is that reverse-mode
AD must retain or reconstruct the computational trace of the segmented
simulation. This can become expensive when the number of states, events, or
solver steps grows. More details can be found in Appendix \ref{app:ad_through_simulation_dae_details}.

\section{Discrete-Adjoint Parameter Learning for DAEs with Events}
\label{sec:discrete_adjoint_dae_events}

The AD-through-simulation approach requires a simulator whose numerical
operations are visible to the automatic-differentiation system. Many mature
DAE solvers, however, are available only as external numerical engines. The
discrete-adjoint approach targets this setting. It uses the external solver to
generate the event-split forward trajectory at the current parameter values,
including the event times and pre- and post-event states. These quantities are
then used to construct a differentiable discrete residual system, from which
the Lagrange multipliers and parameter gradient are computed explicitly.

The discrete-adjoint method starts from the same constrained parameter
estimation problem, but keeps the constraints explicit after discretization.
The forward solve produces an event-split trajectory: smooth segments between
events and event records containing the event time, active event index,
pre-event state, and post-event state.

Let segment $m$ be bounded by event times $\tau_m$ and $\tau_{m+1}$. The
implementation stores normalized time coordinates
\begin{equation}
\nonumber
    0=\eta_{m,0}<\eta_{m,1}<\cdots<\eta_{m,N_m-1}=1,
\end{equation}
and reconstructs physical node times as
\begin{equation}
\nonumber
    t_{m,k}=
    \tau_m+\eta_{m,k}(\tau_{m+1}-\tau_m).
\end{equation}
Thus, event times determine both the segment boundaries and the physical time
steps used in the residuals.

For
\[ \nonumber
    w_{m,k}=
    \begin{bmatrix}
        x_{m,k}\\ z_{m,k}
    \end{bmatrix},
\]
the smooth-step residual is the trapezoidal DAE residual
\begin{equation}
\nonumber
    R_{m,k}
    =
    \begin{bmatrix}
        -x_{m,k+1}
        +x_{m,k}
        +\frac{h_{m,k}}{2}
        \left[
            f(t_{m,k},x_{m,k},z_{m,k},p)
            +
            f(t_{m,k+1},x_{m,k+1},z_{m,k+1},p)
        \right] \\
        g(t_{m,k+1},x_{m,k+1},z_{m,k+1},p)
    \end{bmatrix}
    =
    0,
\end{equation}
\nonumber
where $h_{m,k}=t_{m,k+1}-t_{m,k}$. Event blocks are represented by residuals
of the form
\begin{equation}
\nonumber
    E_m(\tau_m,w_m^-,w_m^+,p)=0,
\end{equation}
which enforce the guard condition, reset equations, continuity of unchanged
variables, and algebraic consistency after the event.

The finite-dimensional constrained problem is therefore
\begin{equation}
\nonumber
\begin{aligned}
    \min_{W,\tau,p} \quad
        & \mathcal{J}_h(W,\tau,p) \\
    \mathrm{s.t.}\quad
        & R_{m,k}(W,\tau,p)=0, \\
        & E_m(W,\tau,p)=0,
\end{aligned}
\label{eq:discrete_constrained_problem_fluent}
\end{equation}
where $W$ denote the collection of the state and algebraic variables.
Problem (\ref{eq:discrete_constrained_problem_fluent}) is not a penalty relaxation. It is the discrete constrained form of the
same parameter-estimation problem. The corresponding discrete Lagrangian is
\begin{equation}
\nonumber
    \mathcal{L}_h
    =
    \mathcal{J}_h(W,\tau,p)
    +
    \sum_{m,k}\lambda_{m,k}^{\top}R_{m,k}(W,\tau,p)
    +
    \sum_m\mu_m^{\top}E_m(W,\tau,p).
\end{equation}
Because the residuals vanish on feasible trajectories,
$\mathcal{L}_h=\mathcal{J}_h$ for any multipliers. The multipliers are chosen
so that the stationarity conditions with respect to $W$ and $\tau$ eliminate
the need to compute explicit trajectory and event-time sensitivities. The
gradient is then obtained from
\begin{equation}
\nonumber
    \frac{d\mathcal{J}_h}{dp}
    =
    \frac{\partial \mathcal{L}_h}{\partial p}.
\end{equation}

Computationally, the gradient is evaluated by a reverse block sweep. Smooth
segment blocks solve local adjoint systems associated with the trapezoidal DAE
residuals. Event blocks solve event-adjoint equations associated with the guard
and reinitialization residuals. Because event times define adjacent segment
boundaries, event-time stationarity couples an event block with the segment
immediately before it. The implementation handles this by storing an affine
pending-event adjoint and resolving its scalar coefficient when the preceding
segment is swept. The discrete-adjoint route trades implementation convenience for control over
the gradient calculation. Its accuracy is tied to the residual discretization
used for the adjoint calculation, not necessarily to the tolerance of the
forward DAE solver. This distinction explains why a high-accuracy forward solve
does not automatically imply a high-accuracy discrete-adjoint gradient: the
adjoint residual mesh and quadrature must be sufficiently accurate for the
optimization problem. Conversely, because the method does not require
differentiating through the solver internals, it can be paired with mature
external DAE solvers and can expose smaller linear systems when symbolic
preprocessing or tearing is available. More details can be found in Appendix \ref{app:discrete_adjoint_details}.

\section{Experiments}
\label{sec:experiments}

We evaluate the proposed gradient pipeline on two hybrid-system parameter
identification problems. The first  is an 
electrical ladder network with a state-dependent reinit event, and is
used to compare the implicit-AD route (diffrax with a JAX-AD-friendly
algebraic solver) against the discrete-adjoint route (Sundials/IDA
forward solve, padded backward sweep for computing the multipliers). The second
is a planar multi-ball system with
contact-driven reinit events; we use it to study how each route scales with the number of balls~$N$ and to position our JAX-AD pipeline against the established PyTorch-AD baseline. The code for generating the experimental results can be found at {\url{https://github.com/ionmatei/diff-dae-events/}.

\subsection{Cauer ladder with a state-dependent event}
\label{sec:exp:cauer}

The Cauer DAE has 7 differential states (capacitor voltages $C_1$--$C_5$
and inductor currents $L_1$, $L_2$), 5 algebraic states, and 7
positive-valued parameters
$\theta=(C_1, C_2, C_3, C_4, C_5, L_1, L_2)$. A single \texttt{when}
clause resets $C_3$ when its voltage crosses $0.5$ V. The trajectory over
20 seconds exhibits $11$ event firings.
We initialize the parameters from the truth via a log-uniform
multiplicative bias of half-width $0.1$ (so each component lies in
$[\theta^\star\!\cdot\! e^{-0.1},\theta^\star\!\cdot\! e^{+0.1}]$,
i.e.\ a $\pm 10\%$ band) and run Adam for $500$ iterations with step
size $10^{-2}$ on $500$ uniform target
points. Both routes use the same sigmoid blending sharpness
$\beta=150$ across event boundaries to widen the loss basin by smoothing the effects of the events (see \ref{sec:blending}). 

Table~\ref{tab:cauer-events} reports the relative error per parameter,
the final loss of the chosen snapshot, and the wall-clock cost. The
implicit-AD route attains a lower final loss than the discrete-adjoint
route ($1.69\times 10^{-6}$ vs. $4.00\times 10^{-4}$). It also recovers
several parameters to below one percent relative error, although the
remaining error on $C_5$ indicates that some parameter directions are
weakly identifiable from the chosen output and time grid. The
discrete-adjoint route recovers $C_3$, $C_4$, and $L_1$ accurately, but
retains larger residual errors on $C_1$, $C_5$, and $L_2$.
Figure \ref{fig:cauer-overlays} overlays the optimized trajectories on the
ground truth, and Figures \ref{fig:cauer-loss} shows the loss histories.

\begin{table}[htp!]
  \centering
  \small
  \caption{Cauer ladder with a state-dependent reinit event. Per-parameter
  relative error $|\theta - \theta^\star| / |\theta^\star|$, expressed
  in percent, after Adam optimization with a $\pm10\%$ initial bias and
  $500$ iterations. Best-loss snapshot reported. AD: implicit
  differentiation through diffrax with a JAX-AD-friendly algebraic
  solver. DA: discrete adjoint via Sundials/IDA forward and padded
  backward sweep.}
  \label{tab:cauer-events}
  \begin{tabular}{lccccccccc}
    \toprule
    Method & $C_1$ & $C_2$ & $C_3$ & $C_4$ & $C_5$ & $L_1$ & $L_2$ &
    Loss & Time/iter \\
    \midrule
    Initial & 2.78 & 4.50 & 8.77 & 9.22 & 6.47 & 8.61 & 2.16 &
    {7.20e-3}  & --- \\
    AD       & {0.29} & 1.84 & {0.01} & 0.09 &
    {3.28} & 1.13 & {0.04} &
    {{1.69e-6}} & 0.03 s \\
    DA       & 2.97 & {1.80} & 0.02 & {0.03} & 5.44 &
    {0.53} & 1.09 &
    {4.00e-4}  & 0.042 s \\
    \bottomrule
  \end{tabular}
\end{table}

\begin{figure}[htp!]
  \centering
  \begin{subfigure}[b]{0.48\linewidth}
    \includegraphics[width=\linewidth]{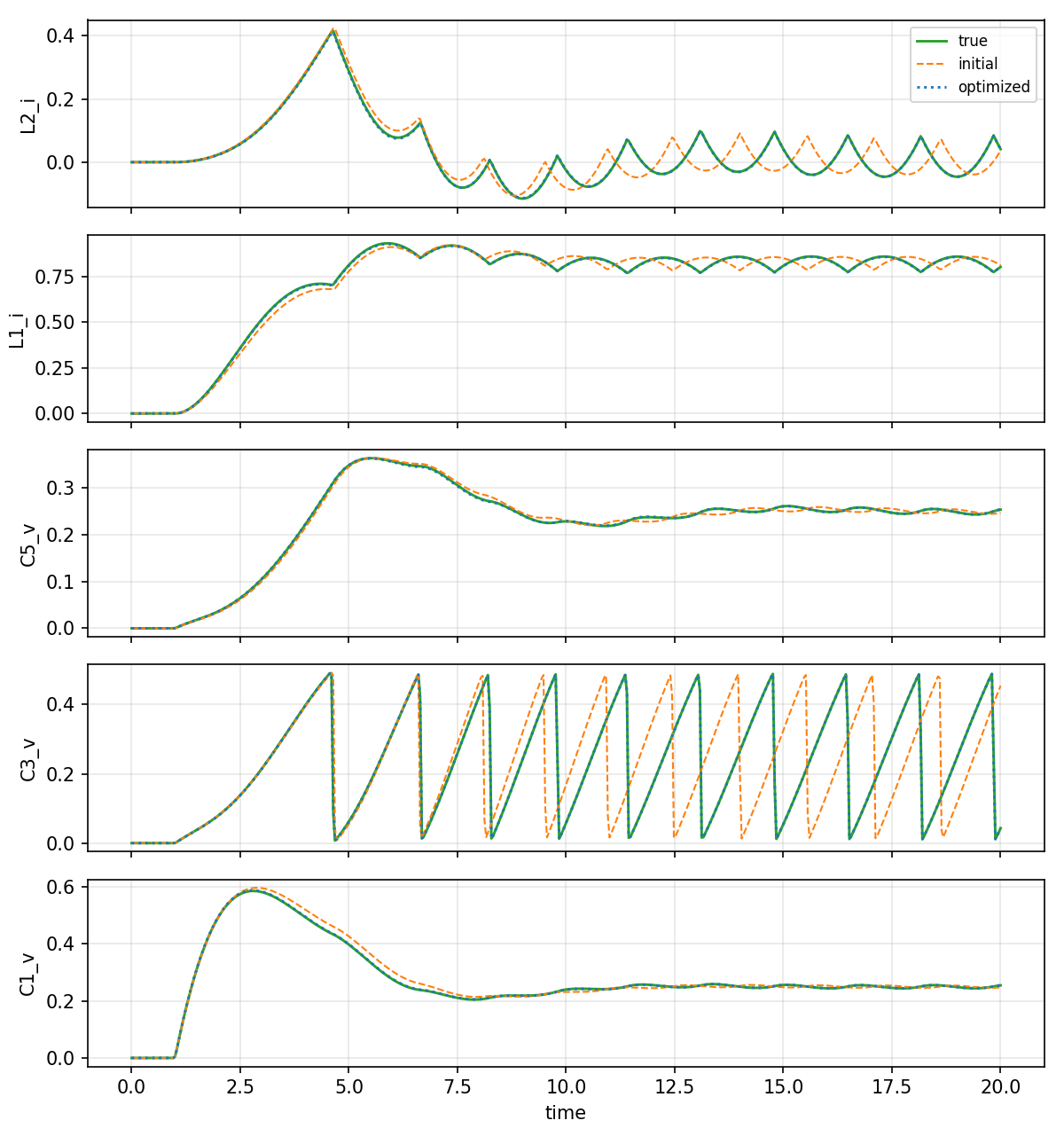}
    \caption{Implicit-AD route.}
  \end{subfigure}
  \hfill
  \begin{subfigure}[b]{0.48\linewidth}
    \includegraphics[width=\linewidth]{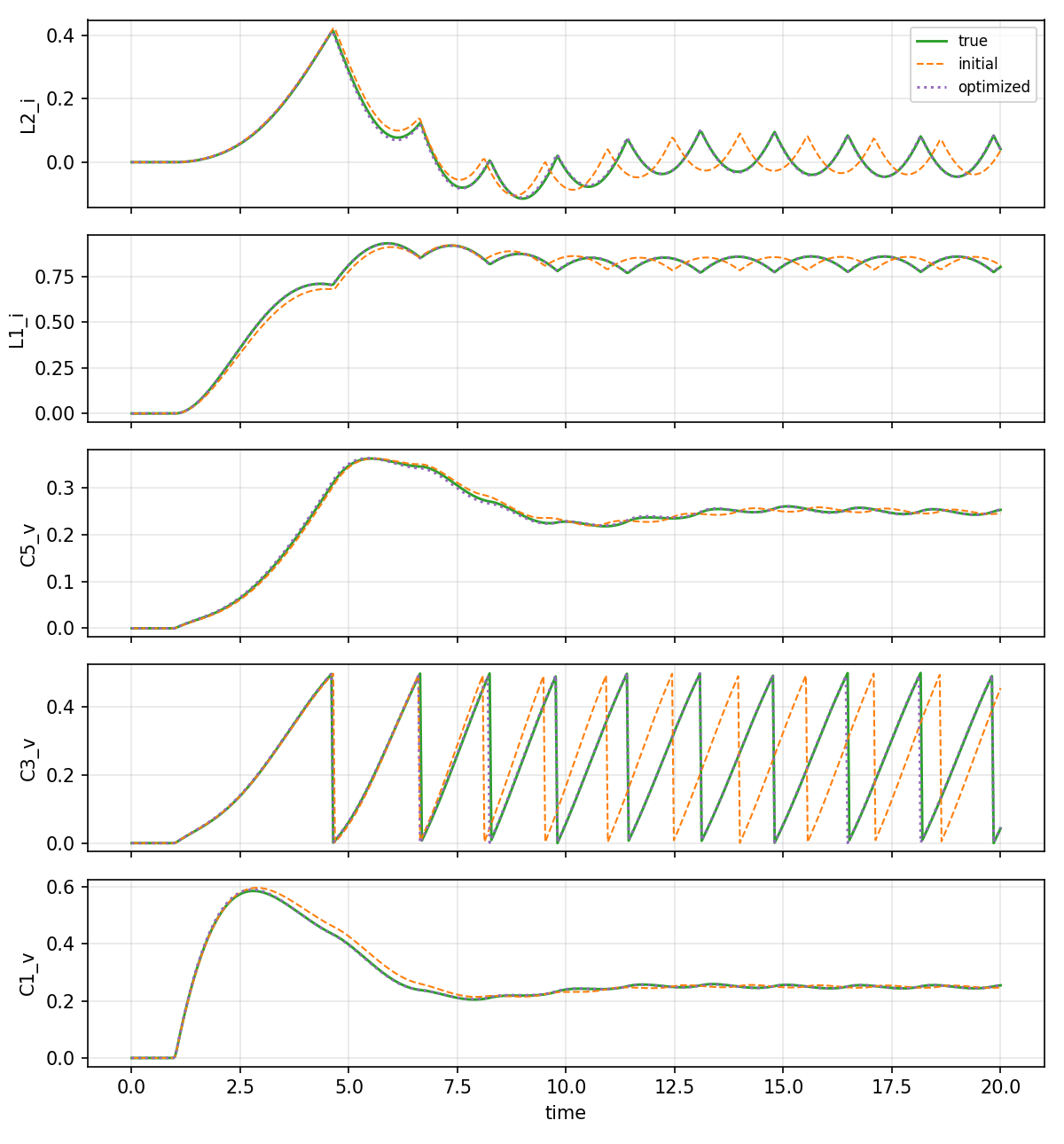}
    \caption{Discrete-adjoint route.}
  \end{subfigure}
  \caption{Cauer ladder: optimized vs.\ true trajectories at the
  best-loss snapshot. Both routes capture the qualitative event
  structure; the AD route additionally matches the post-reset
  oscillation envelope.}
  \label{fig:cauer-overlays}
\end{figure}

\begin{figure}[t]
  \centering
  \begin{subfigure}[b]{0.48\linewidth}
    \includegraphics[width=\linewidth]{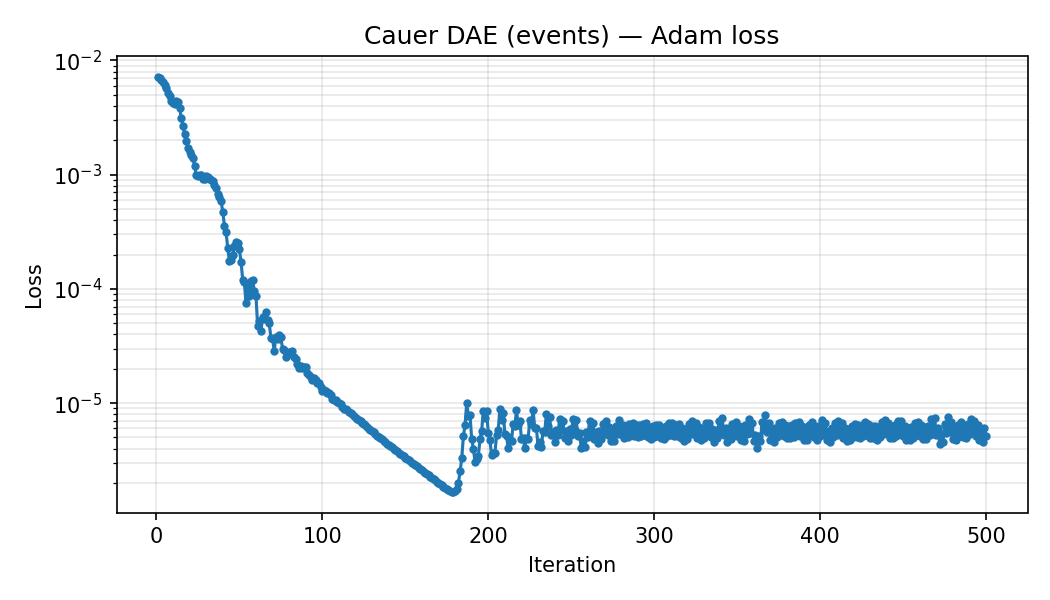}
    \caption{AD loss history.}
  \end{subfigure}
  \hfill
  \begin{subfigure}[b]{0.48\linewidth}
    \includegraphics[width=\linewidth]{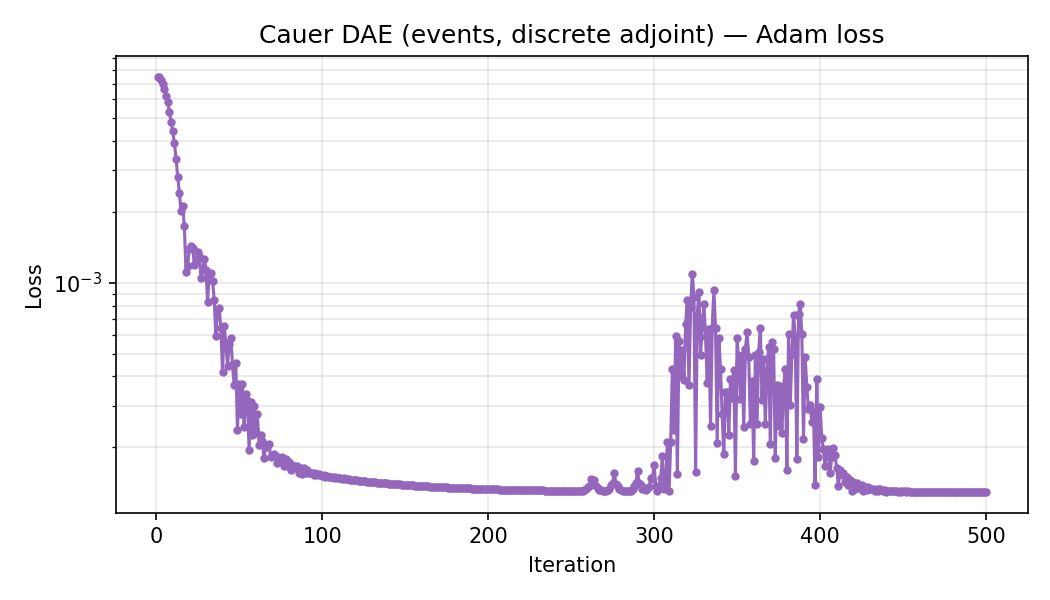}
    \caption{DA loss history.}
  \end{subfigure}
  \caption{Cauer ladder: Adam loss histories on log scale. The AD
  curve descends $\sim$5 decades whereas the DA curve plateaus at
  $\sim 10^{-4}$; the gap is consistent across the last
  $\sim 100$ iterations.}
  \label{fig:cauer-loss}
\end{figure}

The parameter errors should be interpreted in the context of output
sensitivity and identifiability. Different parameter combinations can produce nearly indistinguishable output
trajectories, especially along weakly identifiable directions. Therefore,
matching the output loss is the primary optimization objective, while exact
parameter recovery depends on the informativeness of the measured outputs and
the time grid.

From the optimization perspective any parameter combination that minimize the loss is acceptable (i.e., there may be multiple local minima). The accuracy gap between the two routes on this benchmark traces back
to two effects. First, the discrete-adjoint pipeline constructs the Jacobian based on a trapezoidal-step approach, so its gradient quality is bounded by the
adjoint mesh density rather than by the forward solver tolerance.
Second, the loss surface near the truth is anisotropic and dominated
by long-time-scale plateaus between events; with a uniform target
grid the trapezoidal Jacobians of the discrete adjoint accumulate a
small but coherent bias along the soft directions, observable here as
the residual error on $C_5$ and $C_1$. The implicit-AD pipeline gradient inherits the diffrax tolerance
($\mathrm{rtol}{=}\mathrm{atol}{=}10^{-6}$) directly. The price is
that AD must hold the entire diffrax tape per segment in JAX memory;
on this 7-parameter system this is negligible, but it scales with the
number of state dimensions and the length of the dense-output tape, a
distinction that drives the next experiment. The reported loss function values are computed based on the best parameters without the segment blending that has a smoothing effect.

\subsection{Multi-ball planar bouncing system: scaling with $N$}
\label{sec:exp:balls}
We consider a planar system of $N$ identical bouncing balls in a square
box of half-width $10$, with elastic collisions against the four walls
(restitution $e_b$) and inelastic ground impacts (restitution $e_g$).
The dynamics is an ODE with state-dependent events, $4N+\binom{N}{2}$
event clauses for $N\in\{3,7,15\}$. We optimize three identifiable
parameters $\theta=(g, e_g, e_b)$ from a randomly biased start
($\pm 10\%$ log-uniform). All three identification routes share the
same $500$ points uniform target grid, the same Adam step
size, and the same convergence tolerance; the only differences are
how the gradient is obtained and which framework holds the autograd
tape:
\begin{description}
  \setlength{\itemsep}{2pt}
  \item[\texttt{jax\_ad}] our JAX/diffrax pipeline with implicit AD,
  same algebraic-solver custom-jvp as in \ref{sec:exp:cauer}.
  \item[\texttt{jax\_da}] the JAX padded-discrete-adjoint pipeline of
  \ref{sec:exp:cauer}, ported to ODE-with-events.
  \item[\texttt{pytorch\_ad}] PyTorch baseline using \texttt{torchdiffeq}
  with hand-written event-time root-finding, AD through the entire
  segmented integration tape.
\end{description}

Table \ref{tab:balls-scaling} summarizes the result. Per-parameter
relative errors are reported as percentages, the final validation
loss is on a denser held-out target grid, and timing is total
wall-clock for the budget of iterations actually used.

\begin{table}[t]
  \centering
  \small
  \caption{Multi-ball bouncing system with $N\in\{3,7,15\}$ balls.
  Per-parameter relative error in \%, final validation loss, and
  total wall-clock time. Budget: $300$ iters for $N\in\{3,7\}$ and
  $600$ iters for $N=15$. Times
  measured on a single CPU thread.}
  \label{tab:balls-scaling}
  \begin{tabular}{llrrrrrrr}
    \toprule
    $N$ & Method & $|\Delta g|/g$ & $|\Delta e_g|/e_g$ & $|\Delta e_b|/e_b$ &
    Loss & Iter & Total time & ms/it \\
    \midrule
    \multirow{3}{*}{3}
      & \texttt{jax\_ad}     & \textbf{0.002} & \textbf{0.000} &
        \textbf{0.001} & \textbf{{1.40e-9}} & 300 & {33.4}{sec}
        & 73.9 \\
      & \texttt{jax\_da}     & 0.27 & 1.16 & 0.16 & {9.38e-5} & 300
        & {22.1}{sec} & 61.9 \\
      & \texttt{pytorch\_ad} & 0.003 & 0.001 & 0.003 & {3.89e-9} & 300
        & {168.6}{sec} & 554.4 \\
    \midrule
    \multirow{3}{*}{7}
      & \texttt{jax\_ad}     & \textbf{0.002} & \textbf{0.001} &
        \textbf{0.001} & \textbf{{1.63e-9}} & 300 & {33.4}{sec}
        & 71.2 \\
      & \texttt{jax\_da}     & 0.15 & 0.09 & 0.10 & {7.76e-6} & 300
        & {84.0}{sec} & 254.8 \\
      & \texttt{pytorch\_ad} & 0.013 & 0.010 & 0.010 & {6.61e-8} & 300
        & {525.5}{sec} & 1740.4 \\
    \midrule
    \multirow{3}{*}{15}
      & \texttt{jax\_ad}     & 0.07 & 0.19 & 0.42 & {1.35e-5} & 600
        & {77.2}{sec} & 110.1 \\
      & \texttt{jax\_da}     & 0.21 & 2.67 & 4.04 & {9.28e-4} & 495
         & {595.7}{sec} & 1157.5 \\
      & \texttt{pytorch\_ad} & \textbf{0.000} & \textbf{0.005} &
        \textbf{0.001} & \textbf{{3.88e-8}} & 423 
        & {2826.9}{sec} & 6652.2 \\
    \bottomrule
  \end{tabular}
\end{table}

\begin{figure}[htp!]
  \centering
  \begin{subfigure}[b]{0.6\linewidth}
    \includegraphics[width=\linewidth]{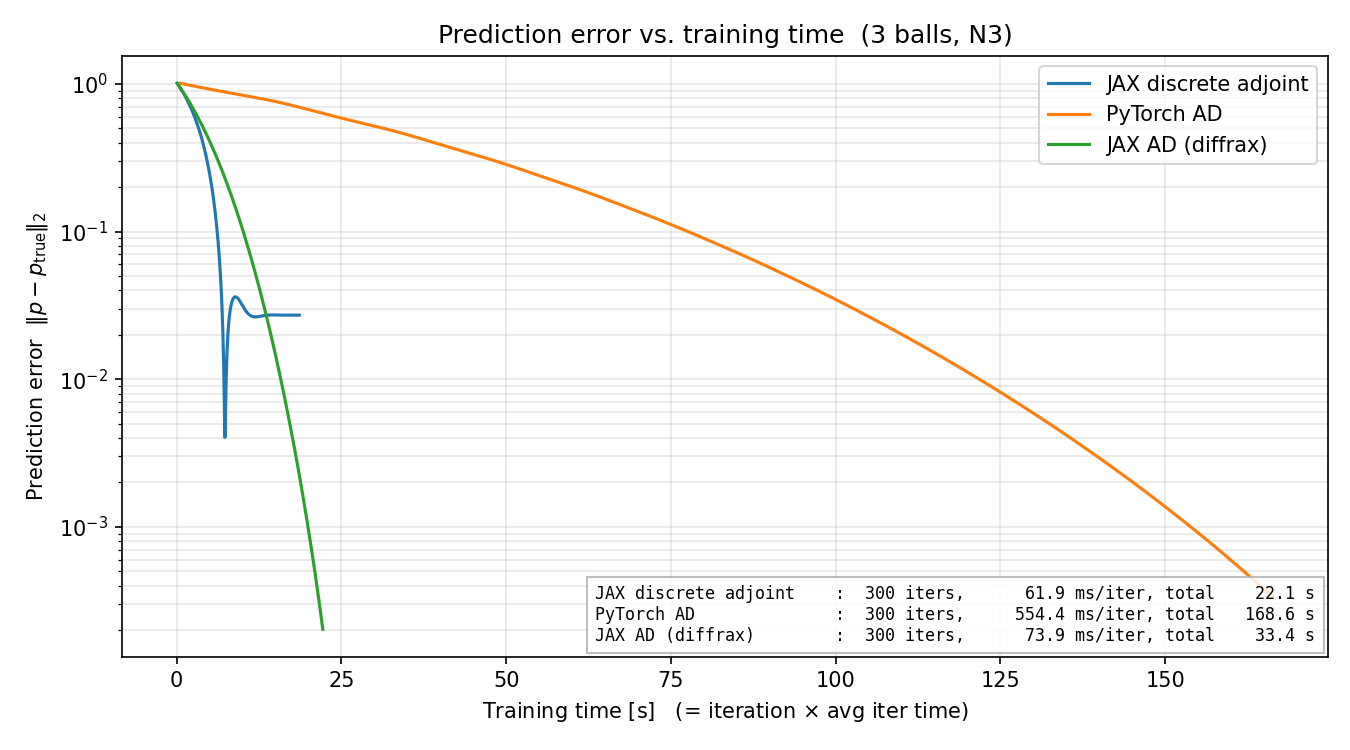}
    \caption{$N=3$.}
  \end{subfigure}
  \hfill \\
  \begin{subfigure}[b]{0.6\linewidth}
    \includegraphics[width=\linewidth]{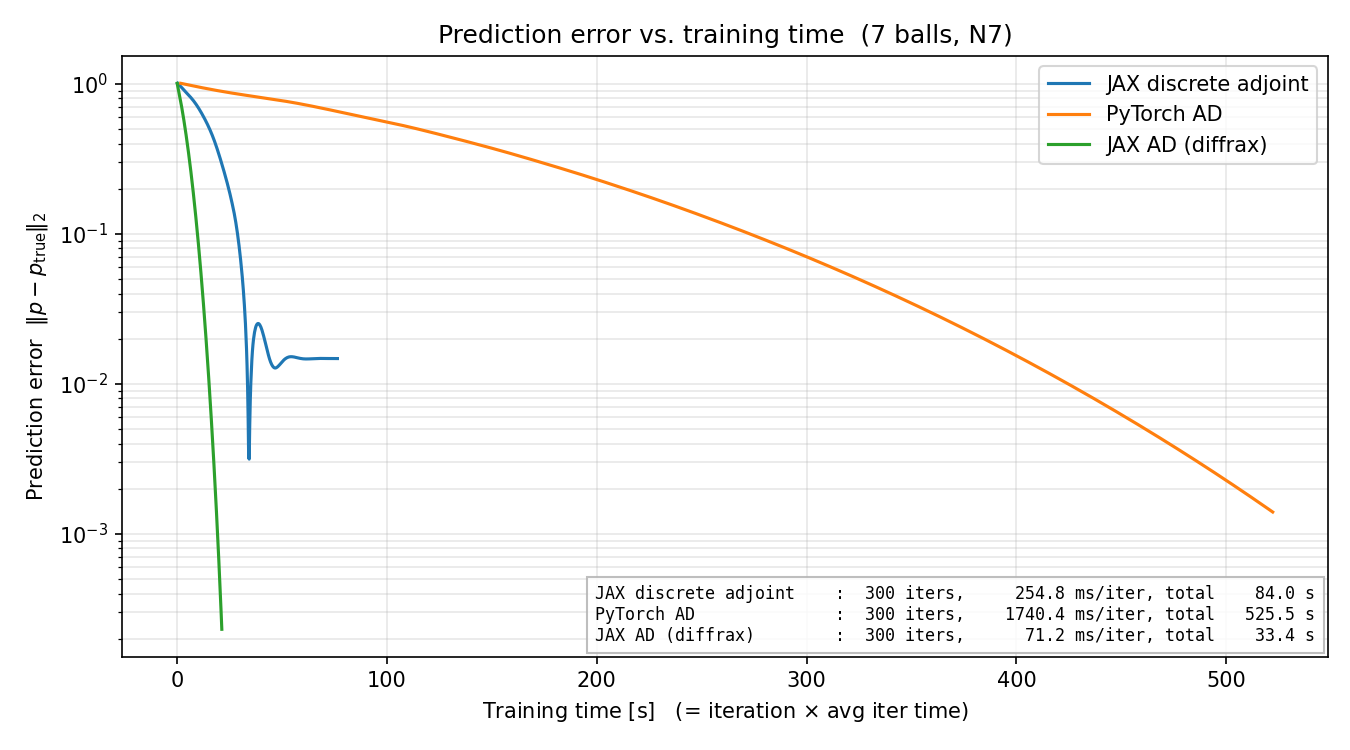}
    \caption{$N=7$.}
  \end{subfigure}
  \hfill \\
  \begin{subfigure}[b]{0.6\linewidth}
    \includegraphics[width=\linewidth]{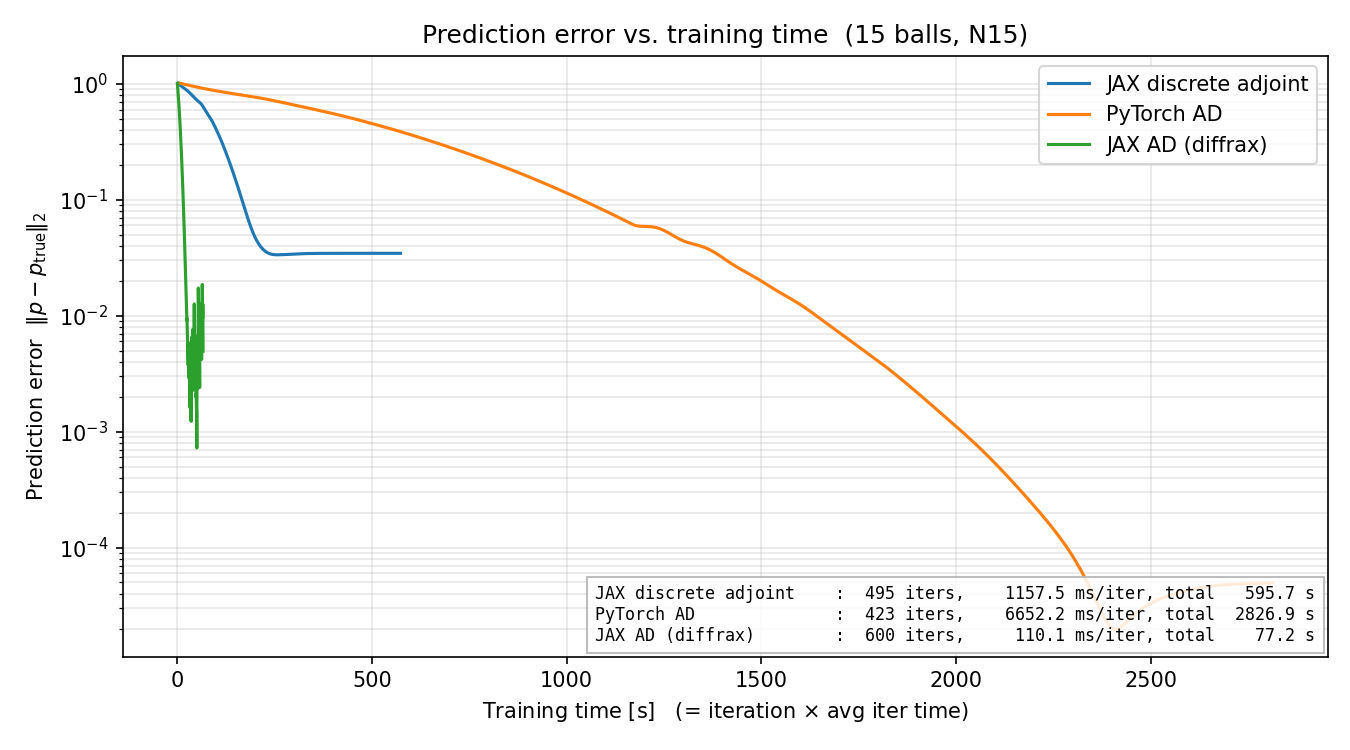}
    \caption{$N=15$.}
  \end{subfigure}
  \caption{Bouncing balls: prediction error vs.\ wall-clock time. The
  horizontal axis exposes the throughput differential: the JAX-AD curve
  reaches the same accuracy plateau as the PyTorch-AD curve in
  $5\!\times$ to $35\!\times$ less wall-clock time across the three
  problem sizes.}
  \label{fig:balls-time}
\end{figure}

Three observations stand out.

\emph{(i) Per-iteration scaling with $N$.} The JAX-AD pipeline is the
flattest: $74 \!\to\! 71 \!\to\! 110$~ms/iter as $N$ goes from $3$ to
$15$. JIT compilation amortizes across iterations, so the marginal
cost is dominated by the diffrax forward and the segment-wise solve;
both stay vectorized in $N$. The PyTorch-AD baseline scales steeply
($554 \!\to\! 1740 \!\to\! 6652$~ms/iter, roughly $3\times$ per
$N\!\to\!2N$), reflecting the unfused per-event-segment work and the
overhead of \texttt{torchdiffeq}'s adaptive stepper rebuilding its
graph each call. The JAX-DA pipeline sits between the two and
inherits the discrete-adjoint cost of building per-step Jacobians,
which scales with the segment count and hence with the event density.

\emph{(ii) Accuracy/throughput frontier.} At $N\!\in\!\{3,7\}$, our
JAX-AD pipeline strictly dominates: it reaches a lower loss (by
$\sim\!2\!\times$) in $\sim\!5\!\times\text{--}16\!\times$ less wall
time. At $N=15$ the comparison flips: PyTorch-AD reaches
$3.9\!\times\!10^{-8}$ versus JAX-AD's $1.4\!\times\!10^{-5}$ but at
$37\!\times$ the cost. Whether the extra accuracy is worth it depends
on the downstream task; for parameter identification under noisy
measurements, the JAX-AD result is well within the noise floor of any
realistic measurement model.

\emph{(iii) When the discrete adjoint trails.} The JAX-DA results
mirror the Cauer experiment: the route is competitive on small
problems but the gradient bias induced by the per-segment
trapezoidal mesh and the uniform target grid grows with $N$, ending
at $\sim\!4\%$ on $e_b$ at $N=15$. This is not a fundamental property of
the discrete adjoint, finer per-segment meshes would close the
gap, but it is a property of the padded fixed-shape JIT regime
that makes the route fast on small problems.

\section{Discussion and Comparison with Existing Approaches}
\label{sec:comparison_sota}

Most differentiable continuous-time learning methods are formulated for ODEs.
Neural ODEs introduced the view of a neural network layer as the solution of an
initial-value problem and popularized adjoint sensitivity methods for
memory-efficient training~\citep{chen2018neuralode}. Implementations such as
\texttt{torchdiffeq} provide differentiable ODE solvers, support direct
backpropagation through solver operations, and expose an adjoint interface for
constant-memory reverse-mode differentiation~\citep{torchdiffeq}. These tools
are effective when the model is an ODE and the state evolves continuously
between observation times.

Event-aware ODE methods extend this setting by adding event functions and reset
maps. They allow models such as bouncing balls or switching systems to be
represented as smooth ODE flows separated by event-triggered state updates
\citep{chen2021neural_event_functions}. In such methods, gradients can pass
through event times, terminal states, and reset maps, provided the event
sequence is fixed locally. This is closely related to our
AD-through-simulation approach. The key difference is that our setting includes
algebraic constraints: the simulator must solve for algebraic variables during
integration, after events, and during output reconstruction.

Other recent work uses ODEs to model irregular event sequences. For example,
COTODE constructs continuous latent trajectories for event-sequence modeling by
interpolating event embeddings and integrating a Neural ODE-type model
\citep{kuleshov2024cotode}. This addresses a different problem. In such
models, events are observations or inputs to a latent representation. In our
setting, events are physical or logical constraints generated by guard
equations; they trigger reinitialization rules that must remain consistent with
algebraic equations.

Classical DAE solvers and equation-based simulation tools already provide
robust mechanisms for forward simulation of DAEs with events, including root
finding, state reinitialization, and consistency restoration. However, they are
not usually designed for end-to-end differentiable parameter learning.
Sensitivity analysis for DAEs is well established, but events add additional
terms through event-time shifts and reset maps. Practical simulation tools also
often expose the forward trajectory without exposing the residual-level
structure needed for efficient gradient-based optimization.

Our work occupies the intersection of these directions. Compared with
differentiable ODE solvers, we treat algebraic variables as implicit functions
of time, differential states, and parameters, and differentiate them through
the implicit function theorem. Compared with event-sequence Neural ODE models,
we treat events as constraints of the physical model rather than as data
tokens. Compared with classical DAE simulation, we focus on differentiable
parameter learning and provide both an AD-through-simulation implementation and
an explicit discrete-adjoint formulation. Table~\ref{tab:sota_comparison} summarizes the conceptual differences.

\begin{table}[t]
\centering
\caption{Comparison of differentiable event-learning approaches.}
\label{tab:sota_comparison}
\begin{tabularx}{\textwidth}{p{0.23\textwidth}p{0.23\textwidth}p{0.24\textwidth}p{0.20\textwidth}}
\toprule
Approach & Model class & Event treatment & Gradient mechanism \\
\midrule
Neural ODE / differentiable ODE solvers
&
ODEs
&
Usually continuous trajectories; event handling available in some solvers
&
Backpropagation through solver operations or continuous adjoint
\\

Neural event ODEs
&
ODEs with event functions and reset maps
&
Events terminate integration and trigger state updates
&
AD through event time, terminal state, and reset map; event sequence fixed locally
\\

ODE-based event-sequence models
&
Latent ODEs for irregular event data
&
Events are observations or inputs used to construct latent trajectories
&
AD through latent continuous-time model
\\

AD-through-simulation DAE method
&
Semi-explicit DAEs with state-dependent events
&
Events are guard-triggered resets; algebraic variables are re-solved after events
&
AD through reduced DAE simulation, implicit algebraic solve, event time, and reset map
\\

Explicit discrete-adjoint DAE method
&
Discretized hybrid DAE residual system
&
Smooth-segment residuals coupled with event residuals
&
Backward solve for residual Lagrange multipliers; event-time stationarity handled explicitly
\\
\bottomrule
\end{tabularx}
\end{table}

The AD-through-simulation method is closest in spirit to differentiable ODE
solvers with event handling. Its advantage is implementation simplicity: once
the DAE is reduced by solving the algebraic equations, the event-driven
simulation can be written as a differentiable program. Its limitation is that
the returned gradient is the derivative of the computational path selected by
the forward event solver.

The explicit discrete-adjoint method is closer to constrained optimization. It
keeps the residual equations visible and solves for their Lagrange multipliers.
This gives direct control over the discrete residuals, event blocks, and
event-time stationarity conditions. It also clarifies the theoretical role of
residuals: they are not loss penalties, but equality constraints whose
multipliers produce the adjoint gradient.

The discrete-adjoint construction used here is closely related to earlier work
in the control and sensitivity-analysis communities. In power-system and hybrid
systems analysis, trajectory sensitivities for differential-algebraic-discrete
models were developed with explicit jump conditions at switching and reset
events~\citep{galan1999parametric,hiskens2000trajectory}. Adjoint sensitivity
methods were also developed for DAEs, including the derivation and numerical
solution of adjoint DAE systems~\citep{cao2003adjoint}. More recently,
discrete-adjoint methods have been implemented for hybrid dynamical systems
with state- and time-dependent switching, with applications to preventive and
corrective control~\citep{zhang2017discrete}. Our formulation follows this
same first-discretize-then-adjoint philosophy, but focuses on parameter learning
for DAEs with event-triggered reinitialization and makes the event-split
residual structure explicit. he distinction in this paper is not the existence of adjoint methods for
hybrid systems, but the combination of an event-split DAE residual construction
with a practical parameter-learning workflow and a comparison against
AD-through-simulation implementations.

The algebraic solve used in our AD-through-simulation implementation should be
viewed as a baseline realization of implicit differentiation rather than as a
state-of-the-art Modelica-like compilation strategy \citep{fritzson2014principles}. We solve the algebraic equations
with a Newton-type iteration and differentiate the solution map using the
implicit function theorem. This is mathematically standard and sufficient for
exposing the role of algebraic constraints in gradient computation. However,
production DAE solvers and Modelica tools typically perform substantial symbolic
and structural preprocessing before any numerical nonlinear solve is called.
Classical DAE solvers such as DASSL and the SUNDIALS family provide robust
implicit integration and nonlinear-solver infrastructure for DAE/ODE systems
\citep{hindmarsh2005sundials,petzold1982dassl}. In Modelica-based workflows,
the compiler first analyzes the equation structure, performs matching and
sorting, identifies algebraic loops, and applies tearing to reduce the number
of variables that must be solved simultaneously
\citep{casella2015modelica,elmqvist1994tearing,tauber2014tearing,zimmer2009module}.
Thus, our Newton/implicit-function approach should be understood as a clear
differentiable reference implementation. A production implementation could
replace the full algebraic solve by the smaller torn algebraic loops produced
by a Modelica compiler, thereby reducing both the forward nonlinear solve and
the backward implicit linear solve. Related work on equation-based algorithmic differentiation also emphasizes
that exploiting the equation structure of DAE models can be more memory
efficient than differentiating a fully expanded simulation trace
\citep{elsheikh2015equation}.

Julia and the SciML ecosystem are highly relevant to differentiable simulation.
Julia was designed to combine high-level numerical programming with performance
comparable to traditional compiled languages~\citep{bezanson2017julia}, and
DifferentialEquations.jl provides a feature-rich ecosystem for ODEs, DAEs, and
hybrid differential equations~\citep{rackauckas2017differentialequations}.
Therefore, a Julia/SciML implementation could be faster for some solver-heavy
or AD-heavy workloads, especially when the solver, event handling, and
sensitivity calculation are compiled and specialized together. Our focus on
JAX and PyTorch is motivated instead by their broader adoption in the machine
learning community and the larger surrounding ecosystem of tooling, examples,
GPU workflows, and developer support~\citep{jax2018github,paszke2019pytorch}.
This difference also makes a direct runtime comparison potentially misleading:
performance would depend not only on the programming language, but also on the
chosen solver, event-location strategy, algebraic-loop handling, AD mode,
compiler specialization, batching strategy, and hardware backend. For this
reason, we treat Julia/SciML as an important complementary implementation path,
but not as a direct baseline unless both implementations are engineered to the
same level of solver and compiler optimization.

\section{Conclusion}
\label{sec:conclusion}

This paper studied differentiable parameter optimization for semi-explicit
DAEs with state-dependent events. The central difficulty is that the
parameter-to-output map is shaped not only by the continuous DAE dynamics, but
also by algebraic consistency constraints, parameter-dependent event times, and
event-triggered reset maps. These features make direct application of standard
ODE-based differentiable simulation methods insufficient without additional
care.

We presented two complementary approaches. The first is an
AD-through-simulation method that reduces the DAE by solving the algebraic
variables inside the vector field and differentiates the algebraic solve using
the implicit function theorem. Events are handled through segmented integration,
composite guard detection, reset maps, and right-continuous target-time
evaluation. This approach is relatively simple to implement in modern AD
frameworks and naturally fits JAX/PyTorch-style differentiable programming.

The second approach is an explicit discrete-adjoint method. It represents the
forward simulation as an event-split residual system, with trapezoidal DAE
residuals on smooth segments and event residuals at guard crossings. The
residuals are treated as equality constraints rather than penalty terms. Their
Lagrange multipliers define the discrete adjoint variables, and a backward
event-split sweep computes the parameter gradient while avoiding explicit
trajectory and event-time sensitivity matrices.

The two methods offer different tradeoffs. AD-through-simulation benefits from
the flexibility of differentiable programming and is easier to integrate with
machine-learning toolchains. The explicit discrete-adjoint method does not require an AD-through simulation engine, it gives more
control over the numerical residuals, event-time stationarity conditions, and
linear algebra used in the gradient computation. In both cases, the gradients
are local to the event path selected by the forward simulation and rely on
transversal guard crossings and unchanged event ordering under small parameter
perturbations.

Future work will focus on improving robustness near nonsmooth event transitions,
including grazing contacts, simultaneous events, and event-order changes. A
second direction is to integrate more advanced DAE compilation techniques, such
as algebraic-loop detection and tearing, so that both the forward algebraic
solve and the backward implicit solve operate on smaller systems. Finally,
larger-scale comparisons across AD frameworks, Modelica-based toolchains, and
production DAE solvers would help clarify the practical performance and
scalability limits of differentiable optimization for hybrid DAE models.

\bibliographystyle{plain}
\bibliography{references}

\clearpage
\appendix
\numberwithin{equation}{section}
\section{Details of the AD-through-Simulation DAE Method}
\label{app:ad_through_simulation_dae_details}

This appendix gives the mathematical and implementation details of the
AD-through-simulation method for DAEs with state-dependent events. The method
treats the event-driven simulator as a differentiable program. A loss is
evaluated by simulating the DAE at prescribed observation times, and the
parameter gradient is obtained by automatic differentiation through the
simulation path selected during the forward pass.

\subsection{DAE specification and parameter layout}

The model is specified by differential states
$x(t)\in\mathbb{R}^{n_x}$, algebraic variables
$z(t)\in\mathbb{R}^{n_z}$, parameters
$p\in\mathbb{R}^{n_p}$, and outputs
$y(t)\in\mathbb{R}^{n_y}$. The semi-explicit DAE has the form
\[
    \dot{x}=f(t,x,z,p),
    \qquad
    0=g(t,x,z,p),
    \qquad
    y=h(t,x,z,p),
\]
where $f$ is the differential-state vector field, $g$ is the algebraic
constraint, and $h$ is the output map.

Only a subset of the full parameter vector may be optimized. Let
$p_{\mathrm{opt}}\in\mathbb{R}^{n_{\mathrm{opt}}}$ denote the optimized
parameter vector, and let $I_{\mathrm{opt}}$ denote the corresponding indices
inside the full parameter vector $p$. During loss evaluation, the full
parameter vector is reconstructed as
\[
    p
    =
    \operatorname{insert}
    \left(
        p_{\mathrm{base}},
        I_{\mathrm{opt}},
        p_{\mathrm{opt}}
    \right),
\]
where $p_{\mathrm{base}}$ contains the fixed baseline values for all parameters
not being optimized. If no explicit output map $h$ is specified, the
differential state $x$ is used as the output.

The equations defining $f$, $g$, $h$, and the event guards are compiled into
JAX-compatible numerical functions. This avoids per-equation Python evaluation
inside the JIT-compiled simulation and gives fixed-shape functions of
$(t,x,z,p)$.

\subsection{Algebraic variables as implicit functions}

The numerical ODE solver advances only the differential state $x$. At each
vector-field evaluation, the algebraic variable $z$ is obtained by solving
\[
    g(t,x,z,p)=0.
\]
When this algebraic equation has a locally unique solution, we denote it by
\[
    z=\zeta(t,x,p),
    \qquad
    g(t,x,\zeta(t,x,p),p)=0.
\]
The DAE is then integrated through the reduced vector field
\[
    \dot{x}
    =
    F(t,x,p)
    =
    f(t,x,\zeta(t,x,p),p),
\]
where $F$ is the vector field passed to the ODE solver.

In the implementation, $\zeta(t,x,p)$ is computed by a chord-Newton iteration.
Let $z^{(0)}$ denote the initial algebraic guess and let $Q$ denote the number
of chord-Newton iterations. A regularized algebraic Jacobian is formed as
\[
    G_z
    =
    \frac{\partial g}{\partial z}(t,x,z^{(0)},p)
    +
    \epsilon_z I,
\]
where $\epsilon_z>0$ is a small regularization constant and $I$ is the identity
matrix of dimension $n_z$. The iteration is
\[
    z^{(q+1)}
    =
    z^{(q)}
    -
    G_z^{-1}g(t,x,z^{(q)},p),
    \qquad q=0,\ldots,Q-1.
\]
The same factorization of $G_z$ is reused across these chord iterations.

The algebraic solve is differentiated using a custom JVP based on the implicit
function theorem. Let
\[
    g_t=\frac{\partial g}{\partial t},
    \qquad
    g_x=\frac{\partial g}{\partial x},
    \qquad
    g_z=\frac{\partial g}{\partial z},
    \qquad
    g_p=\frac{\partial g}{\partial p},
\]
all evaluated at $(t,x,\zeta(t,x,p),p)$. For perturbations
$(dt,dx,dp)$ in $(t,x,p)$, the corresponding algebraic perturbation is $dz$.
Differentiating
\[
    g(t,x,\zeta(t,x,p),p)=0
\]
gives
\[
    g_z\,dz + g_t\,dt + g_x\,dx + g_p\,dp = 0.
\]
Therefore,
\begin{appequation}
    dz=
    -g_z^{-1}
    \left(
        g_t\,dt + g_x\,dx + g_p\,dp
    \right).
    \label{eq:app_ift_jvp}
\end{appequation}
Equation~\eqref{eq:app_ift_jvp} is the JVP used for the algebraic solve. It
allows gradients to propagate through the algebraic constraint without
differentiating through every Newton iteration.

\subsection{Event guards and reset maps}

Events are specified by guard conditions and reinitialization rules. Let
$n_e$ denote the number of event guards. Each guard is represented as a scalar
function
\[
    \phi_e(t,x,z,p),
    \qquad e=1,\ldots,n_e.
\]
The implementation uses the sign convention
\[
    \phi_e>0
    \quad \text{before the event},\qquad
    \phi_e=0
    \quad \text{at the event},\qquad
    \phi_e<0
    \quad \text{after the event}.
\]
Thus, all events are represented as positive-to-nonpositive crossings. For
example, a condition $a>b$ is rewritten as $\phi=b-a$, while a condition
$a<b$ is rewritten as $\phi=a-b$.

The event vector is
\[
    \Phi(t,x,z,p)
    =
    \begin{bmatrix}
        \phi_1(t,x,z,p) \\
        \vdots \\
        \phi_{n_e}(t,x,z,p)
    \end{bmatrix}.
\]
Since guards may depend on algebraic variables, the event vector used by the
solver is evaluated after solving the algebraic equation:
\[
    \Phi(t,x,p)
    =
    \Phi(t,x,\zeta(t,x,p),p).
\]

Each event type $e$ has an associated reset map
\[
    x^+ = \Psi_e(t,x^-,z^-,p),
\]
where $x^-$ and $z^-$ are the pre-event differential and algebraic variables,
and $x^+$ is the post-event differential state. The implementation supports
direct reinitialization of differential states. Algebraic variables are not
reset directly. Instead, after $x^+$ is computed, the post-event algebraic
variable $z^+$ is recomputed from
\[
    g(t,x^+,z^+,p)=0.
\]

For a fixed event index $e$, the reset map is differentiated as an ordinary
JAX function. Its local differential is
\[
    dx^+
    =
    \Psi_{e,x}\,dx^-
    +
    \Psi_{e,z}\,dz^-
    +
    \Psi_{e,p}\,dp
    +
    \Psi_{e,t}\,dt,
\]
where $\Psi_{e,x}$, $\Psi_{e,z}$, $\Psi_{e,p}$, and $\Psi_{e,t}$ denote the
partial derivatives of $\Psi_e$ with respect to $x^-$, $z^-$, $p$, and $t$,
respectively.

\subsection{Segmented event simulation}

The simulation is represented as a sequence of smooth segments separated by
events. Let $t_s$ and $x_s$ denote the start time and initial differential state
of a segment. The algebraic state at the segment start is
\[
    z_s=\zeta(t_s,x_s,p).
\]
The event vector is evaluated at the segment start, and an active-event mask is
formed:
\[
    \alpha_e
    =
    \mathbf{1}
    \{
        \phi_e(t_s,x_s,z_s,p)>\varepsilon_\phi
    \},
    \qquad e=1,\ldots,n_e.
\]
Here $\alpha_e\in\{0,1\}$ indicates whether guard $e$ is active for event
detection on the current segment, and $\varepsilon_\phi>0$ is a small threshold.
Guards already active at the segment boundary are masked out to avoid immediate
retriggering after a reset.

The active guards are combined into a scalar event condition
\[
    \Phi_{\min}(t,x,p;\alpha)
    =
    \min_{e=1,\ldots,n_e}
    \begin{cases}
        \phi_e(t,x,\zeta(t,x,p),p), & \alpha_e=1, \\
        C, & \alpha_e=0,
    \end{cases}
\]
where $C>0$ is a large constant. The ODE solver advances the reduced system
\[
    \dot{x}=F(t,x,p)
\]
until either the terminal time $T$ is reached or the composite guard
$\Phi_{\min}$ crosses zero. A bracketing root finder is then used to refine the
event time.

If an event is detected at time $\tau$, the active event index is selected as
\[
    e^\star
    =
    \arg\min_{e:\alpha_e=1}
    \phi_e(\tau,x(\tau^-),\zeta(\tau,x(\tau^-),p),p).
\]
Here $x(\tau^-)$ denotes the pre-event differential state. The pre-event
algebraic variable is
\[
    z^-=\zeta(\tau,x(\tau^-),p),
\]
and the reset gives
\[
    x(\tau^+)
    =
    \Psi_{e^\star}(\tau,x(\tau^-),z^-,p).
\]
The next segment starts from the post-event pair $(\tau,x(\tau^+))$.

For JIT compilation, event detection is implemented as a fixed-length scan over
a prescribed maximum number of segments, denoted by $K_{\max}$. Each scan step
records
\[
    (a_k,b_k,x_k,r_k),
    \qquad k=1,\ldots,K_{\max}.
\]
Here $a_k$ is the start time of segment $k$, $b_k$ is the segment end time
(either an event time or the terminal time), $x_k$ is the differential state at
$a_k$, and $r_k\in\{0,1\}$ indicates whether the segment is a real physical
segment or padding. If the terminal time is reached before $K_{\max}$ segments
are used, the remaining scan iterations are degenerate padding segments. If the
scan reaches $K_{\max}$ before the terminal time, the trajectory is marked as
saturated and the loss is made non-finite to avoid optimizing against an
invalid simulation.

\subsection{Target-time evaluation and output reconstruction}
\label{sec:blending}

After event detection, each real segment is reintegrated to evaluate the
trajectory at the observation times. Let
\[
    t_1,\ldots,t_N
\]
denote the observation or target times. For segment $k$ with interval
$[a_k,b_k]$, each target time $t_i$ is clipped into the segment interval:
\[
    \tilde{t}_{k,i}
    =
    \min\{\max\{t_i,a_k\},b_k\}.
\]
The segment is then integrated with output saved at all clipped times, giving
\[
    X_{k,i}\approx x(\tilde{t}_{k,i}).
\]
The array $X_{k,i}$ contains candidate state values for all segment-target
pairs. Under hard selection, only the value associated with the segment that
contains $t_i$ is used.

With right-continuous hard selection, the selected segment for target time
$t_i$ is
\[
    k^\star(i)
    =
    \max
    \left\{
        k:
        r_k=1,\;
        t_i\in[a_k,b_k]
    \right\}.
\]
The predicted differential state is
\[
    \hat{x}(t_i;p)=X_{k^\star(i),i}.
\]
Thus, if $t_i$ is exactly an event time, the target is assigned to the
post-event segment.

The implementation also supports optional soft segment selection. For each
segment-target pair, define
\[
    \omega_{k,i}
    =
    \sigma\!\left(\beta(t_i-a_k)\right)
    \sigma\!\left(\beta(b_k-t_i)\right) r_k,
\]
where $\sigma(s)=1/(1+\exp(-s))$ is the logistic sigmoid and
$\beta>0$ controls the sharpness of the segment-membership transition. The
normalized weights are
\[
    \bar{\omega}_{k,i}
    =
    \frac{\omega_{k,i}}
    {\sum_{\ell=1}^{K_{\max}} \omega_{\ell,i}+\epsilon_\omega},
\]
where $\epsilon_\omega>0$ prevents division by zero. The blended state is
\begin{appequation}
    \hat{x}(t_i;p)
    =
    \sum_{k=1}^{K_{\max}} \bar{\omega}_{k,i}X_{k,i}.
    \label{eq:blending}
\end{appequation}
As $\beta\to\infty$, this approaches hard segment selection. For finite
$\beta$, the predicted state changes more smoothly near event boundaries.

Once $\hat{x}(t_i;p)$ is available, the algebraic variable and output are
reconstructed as
\[
    \hat{z}(t_i;p)
    =
    \zeta(t_i,\hat{x}(t_i;p),p),
\]
and
\[
    \hat{y}(t_i;p)
    =
    h(t_i,\hat{x}(t_i;p),\hat{z}(t_i;p),p).
\]

\subsection{Loss, gradient, and outer optimization}

Let $y_i^{\mathrm{data}}$ denote the measured or reference output at target time
$t_i$. The loss is
\[
    \mathcal{J}(p_{\mathrm{opt}})
    =
    \sum_{i=1}^{N}
    \left\|
        \hat{y}(t_i;p_{\mathrm{opt}})
        -
        y_i^{\mathrm{data}}
    \right\|_2^2,
\]
or the corresponding mean loss. The full loss evaluation reconstructs the full
parameter vector $p$, simulates the event-driven DAE, evaluates
$\hat{y}(t_i;p_{\mathrm{opt}})$ at all target times, and computes the output
mismatch.

The implementation computes
\[
    \left(
        \mathcal{J}(p_{\mathrm{opt}}),
        \nabla_{p_{\mathrm{opt}}}\mathcal{J}(p_{\mathrm{opt}})
    \right)
\]
with a JIT-compiled \texttt{value\_and\_grad} call. Gradients propagate through
the algebraic solve, reduced ODE integration, event-time computation, reset
map, target-time selection, and output reconstruction.

The resulting gradient can be used by a first-order method such as Adam or by a
quasi-Newton method such as L-BFGS-B. In both cases, the optimizer uses the
same JIT-compiled loss and gradient. The optimization driver records the loss,
gradient norm, parameter values, and iteration time, and stops if the loss or
gradient becomes non-finite.

\subsection{Local validity}

The AD-through-simulation gradient is local to the event path selected during
the forward pass. It assumes that small parameter perturbations do not change
the number of events, their active indices, or their ordering. It also assumes
transversal guard crossings. For event $e$ at time $\tau$, this means
\[
    \frac{d}{dt}
    \phi_e(t,x(t),z(t),p)
    \bigg|_{t=\tau^-}
    \neq 0,
\]
where $\tau^-$ denotes the left limit before the event.

If an event becomes grazing, two events become simultaneous, or the event
ordering changes, the reduced simulation map can be nonsmooth. In that case,
the returned gradient should be interpreted as the derivative of the selected
computational path rather than as a globally smooth sensitivity.

\section{Details of the Explicit Discrete-Adjoint Method}
\label{app:discrete_adjoint_details}

This appendix describes the discrete-adjoint method. Unlike the
AD-through-simulation approach, this method keeps the discretized DAE and event
constraints visible and computes the gradient by solving for the corresponding
Lagrange multipliers.

\subsection{Event-split discrete trajectory}

The forward simulation produces an event-split trajectory. Smooth segment $m$
stores discrete values
\[
    \{t_{m,k},x_{m,k},z_{m,k},\dot{x}_{m,k}\}_{k=0}^{N_m-1},
\]
where $t_{m,k}$ is the $k$-th time node in segment $m$,
$x_{m,k}$ is the differential state, $z_{m,k}$ is the algebraic state,
$\dot{x}_{m,k}$ is the state derivative, and $N_m$ is the number of nodes in
the segment.

Each event record stores
\[
    \tau_m,\qquad e_m,\qquad w_m^-,\qquad w_m^+.
\]
Here $\tau_m$ is the $m$-th event time,
$e_m\in\{1,\ldots,n_e\}$ is the index of the event guard that fired, and
$n_e$ is the number of event guards. The variables $w_m^-$ and $w_m^+$ denote
the pre- and post-event DAE variables:
\[
    w_m^-=
    \begin{bmatrix}
        x_m^-\\ z_m^-
    \end{bmatrix},
    \qquad
    w_m^+=
    \begin{bmatrix}
        x_m^+\\ z_m^+
    \end{bmatrix}.
\]
The index $e_m$ selects the event-specific guard, reset equations, and
continuity equations used in the event residual.

For fixed-shape compilation, the event-split trajectory is stored in padded
block arrays. Each block is either padding, a smooth segment, or an event:
\[
    \texttt{0}=\text{padding},\qquad
    \texttt{1}=\text{segment},\qquad
    \texttt{2}=\text{event}.
\]
For a segment bounded by event times $\tau_m$ and $\tau_{m+1}$, physical node
times are reconstructed from normalized coordinates:
\begin{appequation}
    t_{m,k}
    =
    \tau_m+\eta_{m,k}(\tau_{m+1}-\tau_m),
    \qquad
    0\leq \eta_{m,k}\leq 1.
    \label{eq:app_time_reconstruction}
\end{appequation}
Thus, event times determine both the segment boundaries and the physical time
grid used in the residuals.

\subsection{Loss derivatives and adjoint loads}

The discrete loss is evaluated by interpolating the event-split trajectory at
the observation times:
\[
    \mathcal{J}_h(W,\tau,p)
    =
    \frac{1}{N_{\mathrm{obs}}}
    \sum_{i=1}^{N_{\mathrm{obs}}}
    \left\|
        \hat{y}(t_i;W,\tau,p)-y_i^{\mathrm{data}}
    \right\|_2^2.
\]
Here $W$ denotes all discrete differential and algebraic variables over all
segments, $\tau=(\tau_1,\ldots,\tau_M)$ denotes the event times, and $p$ denotes
the model parameters.

Automatic differentiation is used only for this loss-evaluation map, producing
the direct derivatives
\[
    \ell_W=\frac{\partial \mathcal{J}_h}{\partial W},
    \qquad
    \ell_p=\frac{\partial \mathcal{J}_h}{\partial p},
    \qquad
    \ell_{\tau}=\frac{\partial \mathcal{J}_h}{\partial \tau}.
\]
For an individual node $w_{m,k}$, we write
\[
    \ell_{m,k}
    =
    \frac{\partial \mathcal{J}_h}{\partial w_{m,k}}.
\]
The vector $\ell_p$ is the direct parameter derivative of the loss. The vector
$\ell_{\tau}$ contains the direct derivatives of the loss with respect to the
event times.

During the backward sweep, we use the following accumulated quantities. The
symbol $a$ denotes an adjoint load with respect to a node variable $w$. The
symbol $q_p$ denotes the running accumulator for the parameter gradient. The
symbols $q_{t_s}$ and $q_{t_e}$ denote accumulated sensitivities with respect
to the start and end times of a smooth segment. When a segment boundary is an
event time, these boundary-time sensitivities contribute to stationarity with
respect to that event time.

\subsection{Smooth-step residuals}

For one smooth step in segment $m$, define
\[
    w_c=w_{m,k},
    \qquad
    w_n=w_{m,k+1},
    \qquad
    t_c=t_{m,k},
    \qquad
    t_n=t_{m,k+1}.
\]
The subscripts $c$ and $n$ denote the current and next nodes in the forward time
direction. With
\[
    w_c=
    \begin{bmatrix}
        x_c\\ z_c
    \end{bmatrix},
    \qquad
    w_n=
    \begin{bmatrix}
        x_n\\ z_n
    \end{bmatrix},
\]
the trapezoidal DAE residual is
\begin{appequation}
    R_k(w_c,w_n,t_c,t_n,p)
    =
    \begin{bmatrix}
        -x_n+x_c+\frac{t_n-t_c}{2}
        \left[
            f(t_c,x_c,z_c,p)
            +
            f(t_n,x_n,z_n,p)
        \right] \\
        g(t_n,x_n,z_n,p)
    \end{bmatrix}.
    \label{eq:app_trapezoidal_residual}
\end{appequation}
The first block is the trapezoidal residual for the differential state. The
second block enforces the algebraic constraint at the end of the step.

The adjoint sweep uses the Jacobians
\[
    J_n=\frac{\partial R_k}{\partial w_n},
    \qquad
    J_c=\frac{\partial R_k}{\partial w_c},
    \qquad
    J_p=\frac{\partial R_k}{\partial p},
\]
and the time derivatives
\[
    r_{t_c}=\frac{\partial R_k}{\partial t_c},
    \qquad
    r_{t_n}=\frac{\partial R_k}{\partial t_n}.
\]
All derivatives are evaluated at the stored forward trajectory and current
parameter value.

\subsection{Backward sweep through a smooth segment}

Let $a_n$ be the adjoint load entering node $w_n$ from later times. This load
contains the direct loss derivative at $w_n$ and all contributions propagated
backward from future steps and events. The local multiplier $\lambda_k$ for the
step residual $R_k=0$ is found from
\[
    J_n^{\top}\lambda_k=-a_n.
\]
The dimension of $\lambda_k$ matches the dimension of the residual $R_k$.

After solving for $\lambda_k$, the load propagated to the previous node is
\[
    a_c
    =
    \ell_c+J_c^{\top}\lambda_k,
\]
where
\[
    \ell_c=\frac{\partial \mathcal{J}_h}{\partial w_c}.
\]
The parameter-gradient accumulator is updated by
\[
    q_p
    \leftarrow
    q_p+J_p^{\top}\lambda_k.
\]
At the beginning of the complete backward sweep, $q_p$ is initialized with the
direct loss derivative $\ell_p$.

The same multiplier contributes to derivatives with respect to the physical
times $t_c$ and $t_n$. Define
\[
    d_c=r_{t_c}^{\top}\lambda_k,
    \qquad
    d_n=r_{t_n}^{\top}\lambda_k.
\]
These are scalar sensitivities of the Lagrangian contribution
$\lambda_k^\top R_k$ with respect to the current and next node times.

Let $t_s$ and $t_e$ denote the start and end times of the segment. From
\eqref{eq:app_time_reconstruction},
\[
    t_c=t_s+\eta_c(t_e-t_s),
    \qquad
    t_n=t_s+\eta_n(t_e-t_s),
\]
where $\eta_c$ and $\eta_n$ are the normalized coordinates associated with
$t_c$ and $t_n$. Therefore, the step contributes to the segment-boundary
sensitivities as
\[
    q_{t_s}
    \leftarrow
    q_{t_s}
    +
    d_c(1-\eta_c)
    +
    d_n(1-\eta_n),
\]
and
\[
    q_{t_e}
    \leftarrow
    q_{t_e}
    +
    d_c\eta_c
    +
    d_n\eta_n.
\]
This is how event-time dependence enters the smooth-segment adjoint sweep: if
$t_s$ or $t_e$ is an event time, then the corresponding boundary sensitivity
contributes to stationarity with respect to that event time.

\subsection{Event residuals}

At event time $\tau_m$, the active event index is $e_m$. To simplify notation in
this subsection, we write $\tau=\tau_m$ and $e=e_m$. The event residual is
\begin{appequation}
    E(\tau,w^+,w^-,p)
    =
    \begin{bmatrix}
        \phi_e(\tau,w^-,p) \\
        \rho_e(\tau,w^+,w^-,p) \\
        c_e(w^+,w^-) \\
        g(\tau,x^+,z^+,p)
    \end{bmatrix}
    =
    0.
    \label{eq:app_event_residual}
\end{appequation}
Here $\phi_e$ is the active guard function, $\rho_e$ contains the
event-specific reset equations, $c_e$ contains continuity equations for
variables not changed by the event, and the final block enforces post-event
algebraic consistency.

Define the event Jacobians
\[
    A=\frac{\partial E}{\partial w^+},
    \qquad
    B=\frac{\partial E}{\partial w^-},
    \qquad
    C=\frac{\partial E}{\partial p},
    \qquad
    e_{\tau}=\frac{\partial E}{\partial \tau}.
\]
These derivatives are evaluated at the stored event data
$(\tau,w^-,w^+,p)$.

\subsection{Event adjoint and pending-event representation}

When the reverse scan reaches an event block, the post-event segment has already
been processed. Let $a^+$ denote the adjoint load arriving from the post-event
side, i.e., the accumulated derivative with respect to $w^+$. The event
multiplier $\mu$ satisfies
\[
    A^{\top}\mu=-a^+.
\]
This equation enforces stationarity with respect to the post-event variables
$w^+$.

The event time $\tau$ also appears in the event residual and in the neighboring
segment time grids. Therefore, stationarity with respect to $\tau$ must also be
enforced. In the implementation, the solution of the event-adjoint equation is
represented in affine form:
\[
    \mu(c)=\mu_0+cv,
    \qquad
    A^{\top}v=0.
\]
Here $\mu_0$ is a particular solution of $A^\top\mu=-a^+$, $v$ is a null vector
of $A^\top$, and $c$ is a scalar coefficient that will be determined from
event-time stationarity. This affine representation is used because the event
block alone does not determine $c$.

The affine event multiplier induces an affine load on the pre-event variables:
\[
    a^-(c)
    =
    B^{\top}\mu(c)
    =
    a^-_0+ca^-_v,
\]
where
\[
    a^-_0=B^{\top}\mu_0,
    \qquad
    a^-_v=B^{\top}v.
\]
The vector $a^-(c)$ is the terminal adjoint load passed to the segment before
the event.

The event contribution to the parameter-gradient accumulator is
\[
    q_p^E(c)
    =
    C^{\top}\mu(c)
    =
    q_{p,0}^E+cq_{p,v}^E,
\]
where
\[
    q_{p,0}^E=C^{\top}\mu_0,
    \qquad
    q_{p,v}^E=C^{\top}v.
\]
The superscript $E$ denotes an event-block contribution. The subscript $0$
denotes the contribution from the particular multiplier $\mu_0$, while the
subscript $v$ denotes the contribution from the null direction $v$.

The event contribution to stationarity with respect to the event time is
\[
    q_{\tau}^E(c)
    =
    \ell_{\tau,m}
    +
    q_{\tau}^{+}
    +
    e_{\tau}^{\top}\mu(c)
    =
    q_{\tau,0}^E+cq_{\tau,v}^E.
\]
Here $\ell_{\tau,m}$ is the direct derivative of the loss with respect to the
event time $\tau_m$, obtained from $\ell_\tau$. The scalar $q_\tau^+$ is the
boundary-time sensitivity contributed by the already-processed post-event
segment. The term $e_{\tau}^{\top}\mu(c)$ is the event-residual contribution
with respect to $\tau$. Therefore,
\[
    q_{\tau,0}^E
    =
    \ell_{\tau,m}
    +
    q_{\tau}^{+}
    +
    e_{\tau}^{\top}\mu_0,
    \qquad
    q_{\tau,v}^E
    =
    e_{\tau}^{\top}v.
\]

The scalar $c$ is not determined at the event block itself because the segment
before the event also contributes to the same event-time stationarity
condition. The event block is therefore stored as a pending event containing
the affine quantities
\[
    a^-_0,\quad a^-_v,\quad
    q_{p,0}^E,\quad q_{p,v}^E,\quad
    q_{\tau,0}^E,\quad q_{\tau,v}^E.
\]

\subsection{Resolving the pending event}

When the reverse scan reaches the segment before the pending event, the segment
end time is the same event time $\tau_m$. Consequently, that segment contributes
to the same event-time stationarity condition used to determine $c$.

The segment is swept twice. The first sweep uses terminal load $a^-_0$ and the
true loss loads on the segment. It produces three quantities:
\[
    a_{s,0},\qquad q_{p,0}^S,\qquad q_{\tau,0}^S.
\]
Here $a_{s,0}$ is the adjoint load propagated to the start of the segment,
$q_{p,0}^S$ is the parameter-gradient contribution from the segment, and
$q_{\tau,0}^S$ is the segment contribution to the derivative with respect to the
segment end time, which is the event time.

The second sweep uses terminal load $a^-_v$ and zero loss loads. It produces
\[
    a_{s,v},\qquad q_{p,v}^S,\qquad q_{\tau,v}^S.
\]
These quantities measure how the segment contribution changes with the
null-direction component $cv$ of the event multiplier. The superscript $S$
denotes a smooth-segment contribution.

Together, the two sweeps define affine segment quantities:
\[
    a_s(c)=a_{s,0}+ca_{s,v},
\]
\[
    q_p^S(c)=q_{p,0}^S+cq_{p,v}^S,
\]
and
\[
    q_{\tau}^S(c)=q_{\tau,0}^S+cq_{\tau,v}^S.
\]

The event time is an internal variable of the event-split trajectory. Hence the
total derivative of the Lagrangian with respect to this event time must vanish:
\[
    0=q_{\tau}^E(c)+q_{\tau}^S(c).
\]
Substituting the affine forms gives
\[
    0
    =
    q_{\tau,0}^E
    +
    cq_{\tau,v}^E
    +
    q_{\tau,0}^S
    +
    cq_{\tau,v}^S.
\]
Therefore,
\begin{appequation}
    c
    =
    -
    \frac{
        q_{\tau,0}^E+q_{\tau,0}^S
    }{
        q_{\tau,v}^E+q_{\tau,v}^S
    }.
    \label{eq:app_event_scalar_solve}
\end{appequation}
A small numerical regularization may be added to the denominator in the
implementation.

After $c$ is known, the affine quantities are evaluated. The load propagated to
the start of the preceding segment is
\[
    a_s=a_{s,0}+ca_{s,v}.
\]
The parameter-gradient accumulator is updated by
\[
    q_p
    \leftarrow
    q_p
    +
    q_{p,0}^E+cq_{p,v}^E
    +
    q_{p,0}^S+cq_{p,v}^S.
\]
The pending event is cleared, and the reverse scan continues to earlier blocks.

\subsection{Total gradient}

The gradient returned by the discrete-adjoint method has the form
\[
    \nabla_p\mathcal{J}_h
    =
    \ell_p
    +
    \sum_{m,k}
    J_{p,m,k}^{\top}\lambda_{m,k}
    +
    \sum_m
    C_m^{\top}\mu_m.
\]
The first term, $\ell_p$, is the direct loss derivative with respect to the
parameters. The second term accumulates smooth-step residual contributions,
where $J_{p,m,k}$ is the parameter Jacobian of the step residual and
$\lambda_{m,k}$ is the corresponding step multiplier. The third term
accumulates event-residual contributions, where $C_m$ is the parameter Jacobian
of the $m$-th event residual and $\mu_m$ is the corresponding event multiplier.

In the implementation, this expression is accumulated in the running quantity
$q_p$. Event-time effects enter through the boundary-time sensitivities and
through the scalar stationarity solve in \eqref{eq:app_event_scalar_solve};
they do not appear as a separate final term because stationarity with respect
to each event time has already been enforced.

Thus, the method computes the gradient of the residual-constrained discrete
problem by solving the Lagrange multipliers explicitly. The residuals are
enforced as equality constraints, not as loss penalties.

\end{document}